%% file: main.tex
\documentclass[11pt, a4paper, logo, copyright, nonumbering]{kwaiyii}
\usepackage[authoryear, compress, round]{natbib}
\usepackage{dblfloatfix}
\usepackage{caption}
\usepackage{dramatist}
\usepackage{xspace}
\usepackage{pifont} 
\usepackage{multirow}
\usepackage{tcolorbox}
\usepackage{xltabular}
\usepackage{longtable}
\usepackage{graphicx}
\usepackage{hyperref}
\usepackage{placeins}
\usepackage{amsfonts}
\usepackage{amsmath}
\usepackage{amssymb}
\usepackage{lineno}
\usepackage{multirow}
\usepackage{adjustbox}
\usepackage{float} 
\usepackage[bottom]{footmisc}

\usepackage{CJKutf8}
\usepackage{subfigure}
\usepackage{setspace}

\usepackage{dsfont}
\usepackage{array} 
\usepackage{tabularx} 
\usepackage{subfigure} 
\usepackage{xcolor} 
\usepackage{algorithm}
\usepackage{algorithmic}
\usepackage{graphicx}
\usepackage{subcaption}   %

\usepackage{lipsum}  
\usepackage{multicol} 

\usepackage{graphicx}
\usepackage{booktabs}
\usepackage{multirow}

\definecolor{keywordcolor}{rgb}{0.7, 0.1, 0.1}   
\definecolor{tacticcolor}{rgb}{0.0, 0.1, 0.6}    
\definecolor{commentcolor}{rgb}{0.4, 0.4, 0.4}   
\definecolor{symbolcolor}{rgb}{0.0, 0.1, 0.6}    
\definecolor{sortcolor}{rgb}{0.1, 0.5, 0.1}      
\definecolor{attributecolor}{rgb}{0.7, 0.1, 0.1} 

\usepackage{listings}

\usepackage{textcomp}
\usepackage{tabularx}      

\usepackage{pifont}
\newcommand{\cmark}{\ding{51}}  
\newcommand{\xmark}{\ding{55}}  

\usepackage{tcolorbox}
\newtcolorbox{promptbox}[1][]{
    colback=gray!10,      
    colframe=black!50,     
    boxrule=0.5mm,        
    arc=1mm,              
    boxsep=0mm,
    fontupper=\ttfamily\scriptsize,  
    width=\textwidth,     
    title=#1,
fonttitle=\ttfamily\footnotesize\centering,
}

\makeatletter
\def\@BTrule[#1]{%
  \ifx\longtable\undefined
    \let\@BTswitch\@BTnormal
  \else\ifx\hline\LT@hline
    \nobreak
    \let\@BTswitch\@BLTrule
  \else
     \let\@BTswitch\@BTnormal
  \fi\fi
  \global\@thisrulewidth=#1\relax
  \ifnum\@thisruleclass=\tw@\vskip\@aboverulesep\else
  \ifnum\@lastruleclass=\z@\vskip\@aboverulesep\else
  \ifnum\@lastruleclass=\@ne\vskip\doublerulesep\fi\fi\fi
  \@BTswitch}
\makeatother

\addto\extrasenglish{
}

 {\begin{list}{}%
         {\setlength{\leftmargin}{#1}}%
         \item[]%
 }
 {\end{list}}
 
\reportnumber{001} 

\renewcommand{\today}{}

{\centering
\title{RLEP: Reinforcement Learning with Experience Replay for LLM Reasoning}}

{\centering
\author[*]{
 Hongzhi Zhang, Jia Fu, Jingyuan Zhang, Kai Fu, Qi Wang, Fuzheng Zhang, Guorui Zhou
\\
Klear Team, Kuaishou Technology
}
}

\begin{document}
\input{sections/0-abstract}

\begin{CJK*}{UTF8}{gbsn}

\maketitle

\input{sections/1-introduction}

\input{sections/2-related-work}

\input{sections/3-method}

\input{sections/4-experiments}

\input{sections/5-conclusion}

\setcitestyle{numbers}
\bibliography{main}
\setcitestyle{authoryear}

\newpage

\input{sections/appendix}

\newpage

\end{CJK*}
\end{document}

%% file: sections/0-abstract.tex
\begin{abstract}
Reinforcement learning (RL) for large language models is an energy‑intensive endeavor: training can be unstable, and the policy may gradually drift away from its pretrained weights.
We present \emph{RLEP}—\textbf{R}einforcement \textbf{L}earning with \textbf{E}xperience re\textbf{P}lay—a two‑phase framework that first collects verified trajectories and then replays them during subsequent training. At every update step, the policy is optimized on mini‑batches that blend newly generated rollouts with these replayed successes. By replaying high‑quality examples, RLEP steers the model away from fruitless exploration, focuses learning on promising reasoning paths, and delivers both faster convergence and stronger final performance.
On the Qwen2.5-Math‑7B base model, {RLEP} reaches baseline peak accuracy with substantially fewer updates and ultimately surpasses it, improving accuracy on AIME‑2024 from 38.2\% to 39.9\%, on AIME‑2025 from 19.8\% to 22.3\%, and on AMC‑2023 from 77.0\% to 82.2\%. 
Our code, datasets, and checkpoints are publicly available at \url{https://github.com/Kwai-Klear/RLEP} to facilitate reproducibility and further research.

\end{abstract}

%% file: sections/1-introduction.tex
\section{Introduction}
\label{sec:intro}

Large language models (LLMs) have recently made rapid progress in reasoning. OpenAI o1\citep{openai2024o1}, DeepSeek R1\citep{guo2025deepseek}, and Qwen3\citep{qwen3} etc. has established a new paradigm for solving complex problems. Reinforcement learning (RL) is a key driver of this advance: R1 shows that even a simple rule‑based reward can consistently improving reasoning ability.

Reinforcement Learning with Verifiable Rewards (RLVR) is no trivial task, it must simultaneously balance {learning capacity}, {policy stability}, and {exploration ability}. \textbf{Learning capacity} ensures that policy updates absorb the knowledge uncovered during exploration; \textbf{policy stability} keeps gradients within reasonable bounds and have the LLM’s weights remain close to their pre‑training initialization so as to avoid catastrophic forgetting; and \textbf{exploration ability} allows the model to discover informative trajectories that make continued training worthwhile. 
Recent work has pinpointed techniques for achieving this balance: DAPO\citep{yu2025dapo} and DrGRPO \citep{liu2025understanding} introduce a token‑mean objective that strengthens learning over long sequences, \emph{clip-higher} biases the model toward low‑probability positive rewards to prevent collapse of the exploration space, and adopting a  high-entropy token-updating strategy \citep{wang2025beyond} is proposed to improve both efficiency and stability.

Nevertheless, RL training is an energy‑intensive journey. A policy departs from its current parameters, explores reward‑bearing reasoning paths, and incorporates what it learns. As training continues, training instability and weight drift can cause progress to plateau—or even regress—so that most runs eventually saturate at a fixed performance level.

\begin{figure}[!bp]
    \centering
    \includegraphics[width=0.75\linewidth]{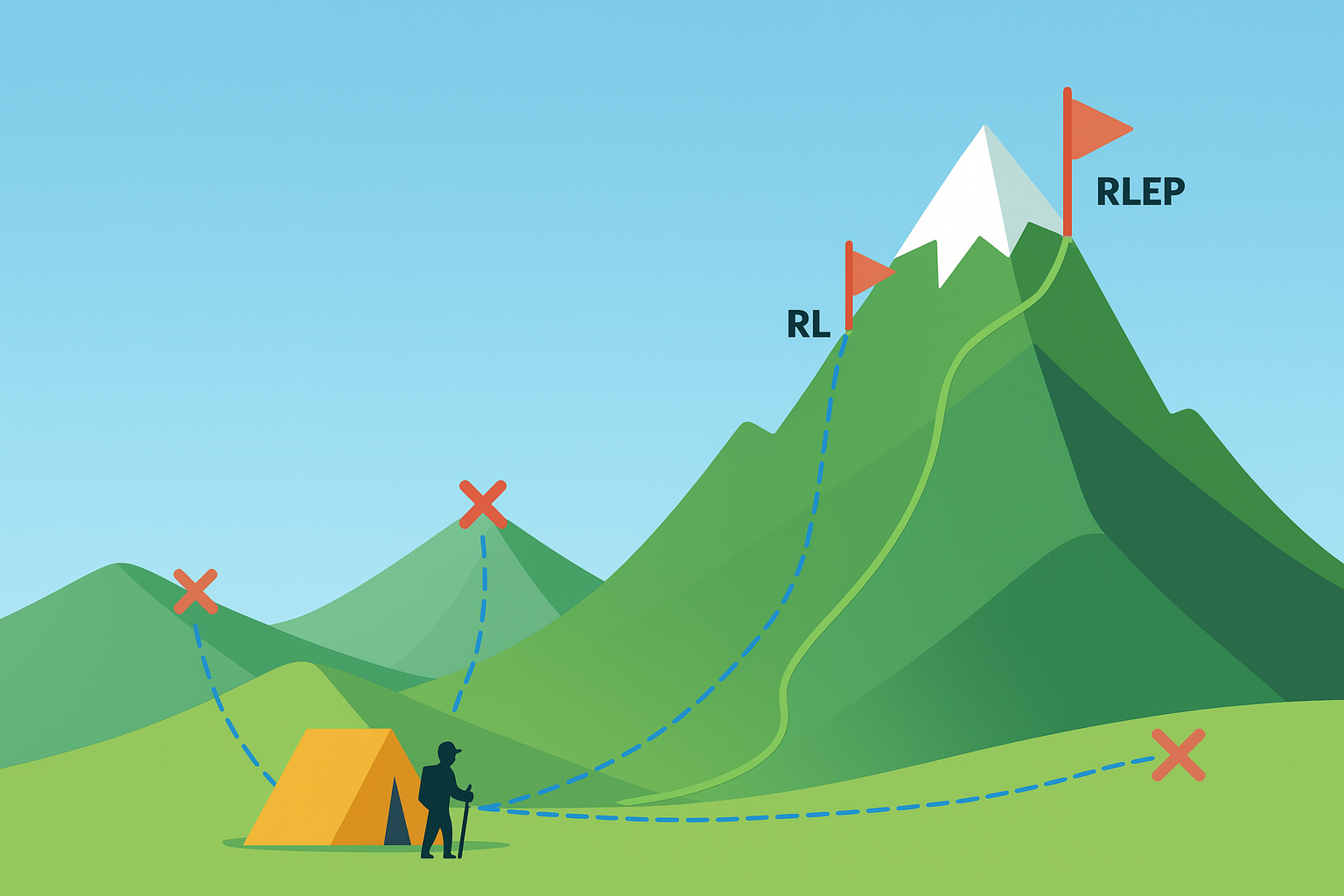}
    \caption{Illustration of reinforcement learning with experience replay.  
During the first trip, the climber explores four candidate routes and reaches the initial red flag, but must stop there due to limited energy.  
With RLEP, the climber quickly replays the successful trajectory to the first flag and then ascends farther to a higher peak.}
    \label{fig:climber_demo}
\end{figure}

The training process is akin to a mountaineering expedition. As Figure \ref{fig:climber_demo} shows, the climber first scouts four possible routes, ultimately reaching the initial red flag—the highest point attainable before exhaustion sets in. On the next journey, the climber retraces those marked waypoints, conserves energy, and pushes onward to a second, higher flag. Motivated by this analogy, we introduce Reinforcement Learning with Experience rePlay (RLEP). By replaying successful trajectories from earlier runs, the policy rapidly recovers its previous best performance at minimal cost and then surpasses it with stable, continued gains. 

RLEP comprises two phases—experience collection and replay‑based training. 
\textbf{Experience collection.} Starting from a model trained with vanilla RL, we generate multiple reasoning trajectories for every question. Trajectories that arrive at the correct answer are retained, forming an experience pool of verified solutions.
\textbf{Replay‑based training. }During each training step, the policy is updated with a mix of freshly generated rollouts and a small batch of successful trajectories sampled from the experience pool. This blend lets the model reinforce proven reasoning chains while still exploring new ones, yielding faster convergence and stronger final performance.

All experiments were performed with the Qwen‑2.5‑7 B model. Experience replay allows the policy to reclaim its previous best score in a few updates and then exceed it latter. Consequently, RLEP achieves stable accuracy gains over the baseline—rising from 38.2 \% to 39.9 \% on AIME 2024 (+1.7 pp), from 19.8 \% to 22.3 \% on AIME 2025 (+2.5 pp), and from 77.0\% to 82.2\% on AMC 2023 (+5.2pp). These results show that experience replay not only accelerates convergence but also produces a higher final performance ceiling.

%% file: sections/2-related-work.tex
\section{Related Work}
\label{sec:related-work}
Experience Replay is a well‑established technique in reinforcement learning that significantly improves sample efficiency and stabilizes training\citep{Lin1992,Mnih2015HumanlevelCT,schaul2015prioritized}. 
\citet{Lin1992} introduced the original ER framework;  \cite{Mnih2015HumanlevelCT} integrated it into Deep Q‑Networks (DQN) and demonstrated its pivotal role in deep RL; and \cite{schaul2015prioritized} proposed Prioritised Experience Replay to further enhance sampling efficiency.

Experience Replay (ER) has recently gained traction in RL training of large language models. 
Existing studies typically deploy replay within the same training run to salvage hard prompts—those for which the current policy cannot yet generate a correct rollout. EFRAME\citep{wang2025eframedeeperreasoningexplorationfilterreplay}, for example, performs extra rollouts on such difficult cases and replays only the trajectories judged valuable, yielding consistent improvements on multimodal tasks. 
The Rollout‑Rescue Mechanism \citep{Polaris2025} adopts a simpler strategy: whenever training encounters a failure, it randomly replaces one incorrect response with a previously buffered correct answer from an earlier epoch. LUFFY \citep{yan2025learningreasonoffpolicyguidance} instead leverages powerful offline guidance like DeepSeek R1 for prompts lacking correct on‑policy rollouts and redesigns advantage estimation to remain valid despite the large policy gap. \cite{dou2025improvingrlexplorationllm} approaches the issue from another angle, using the PPO critic to identify promising states explored early in training and replaying them to preserve exploration capacity. 
Moving beyond these hard‑sample tactics, \cite{bartoldson2025trajectorybalanceasynchronydecoupling} proposes a replay‑buffer‑centred framework that decouples rollout generation from actor updates, simultaneously boosting decoding diversity and training speed.

In contrast, our method gathers trajectories from a converged policy—thereby drawing on inherently stable states—then restarts training from scratch while interleaving these stable trajectories with freshly sampled rollouts. The replay accelerates convergence and smooths learning, whereas the new rollouts safeguard exploration. Crucially, we apply ER uniformly to all prompts rather than restricting it to difficult cases, extending the benefits of replay to the entire training distribution.

%% file: sections/3-Method.tex
\section{RL with Experience Replay}
\label{sec:method}

\subsection{GRPO: Methodology and Recent Advances}

\paragraph{Review of GRPO.}
Following \citet{shao2024deepseekmath} GRPO treats each question $q$ by sampling a \emph{group} of $G$ candidate answers,
$\mathcal{O} \;=\; \{o_1,o_2,\dots,o_G\}$,
from the current policy $\pi_{\theta}$.  
A scalar reward $r_i$ is assigned to every answer $o_i$, yielding a reward vector
\(
\mathcal{R} = \{r_1,r_2,\dots,r_G\}.
\)
The advantage is then used to scale the policy‑gradient update, so that trajectories whose rewards stand above the group mean are reinforced, whereas those below the mean are suppressed:  

\begin{equation}
  A_i \;=\;
  \frac%
      {r_i \;-\; \operatorname{mean}\!\bigl(\{r_1,r_2,\dots,r_G\}\bigr)}%
      {\operatorname{std}\!\bigl(\{r_1,r_2,\dots,r_G\}\bigr)}.
\end{equation}

Let $\pi_{\theta}$ be the current policy and $\pi_{\theta_{\text{old}}}$ the
behaviour policy from the previous iteration.  Omitting the
KL‑divergence regulariser, GRPO updates $\theta$ by maximising
the following objective:
\begin{equation}
\begin{aligned}
\mathcal{J}_{\text{GRPO}}(\theta)=
\mathbb{E}_{\begin{subarray}{r}
  q \sim P(Q),\\
  \{o_i\}_{i=1}^{G}\sim\pi_{\theta_{\text{old}}}(O\,|\,q)
\end{subarray}}
\biggl[ 
\frac{1}{G}\sum_{i=1}^{G}
\bigl(
  \min\!\bigl(
    \tfrac{\pi_{\theta}(o_i\,|\,q)}{\pi_{\theta_{\text{old}}}(o_i\,|\,q)}A_i,\;
    \operatorname{clip}\bigl(
      \tfrac{\pi_{\theta}(o_i\,|\,q)}{\pi_{\theta_{\text{old}}}(o_i\,|\,q)},
      1-\varepsilon,\;1+\varepsilon
    \bigr) A_i
  \bigr)
\bigr)
\biggr]
\end{aligned}
\end{equation}

\paragraph{Enhancements: \emph{token‑mean} and \emph{clip‑higher}.}
Two refinements have been proposed to stabilise GRPO training:
\begin{enumerate}[label=(\alph*)]
  \item \textbf{Token‑mean}~\cite{liu2025understanding,yu2025dapo}.  
        Instead of taking a \emph{sequence}‑level average and then performing a macro‑average over the $G$ answers, token‑mean averages the log‑probability ratios \emph{token by token}.  
        This prevents long, erroneous sequences from being under‑penalised and preserves the learning signal for long, correct sequences.

  \item \textbf{Clip‑higher}~\cite{yu2025dapo}.  
        Positive-advantage trajectories are clipped with a \emph{higher} upper bound than the standard PPO limit, while negative-advantage trajectories keep the usual lower bound.  
This asymmetric clipping mitigates entropy collapse during RL training and plays a key role in balancing exploitation with continued exploration.
\end{enumerate}

With these two strategies, the GRPO objective is revised to

\begin{equation}
\begin{aligned}
\mathcal{J}_{\text{GRPO+}}(\theta)=
\mathbb{E}_{\substack{q \sim P(Q),\\ \{o_i\}_{i=1}^{G}\sim\pi_{\theta_{\text{old}}}(O\,|\,q)}}
\left[
\frac{1}{\sum_{i=1}^{G} |o_i|}
\sum_{i=1}^{G}
\sum_{t=1}^{|o_i|}
\min\left(
r_{i,t}(\theta)\,{A}_{i,t},\;
\operatorname{clip}\left(
r_{i,t}(\theta),\;1{-}\varepsilon_{\text{low}},\;1{+}\varepsilon_{\text{high}}
\right)A_{i,t}
\right)
\right]
\end{aligned}
\end{equation}

\subsection{RL traning with Experience Replay}

\begin{figure}[!h]
    \centering
    \includegraphics[width=0.98\linewidth]{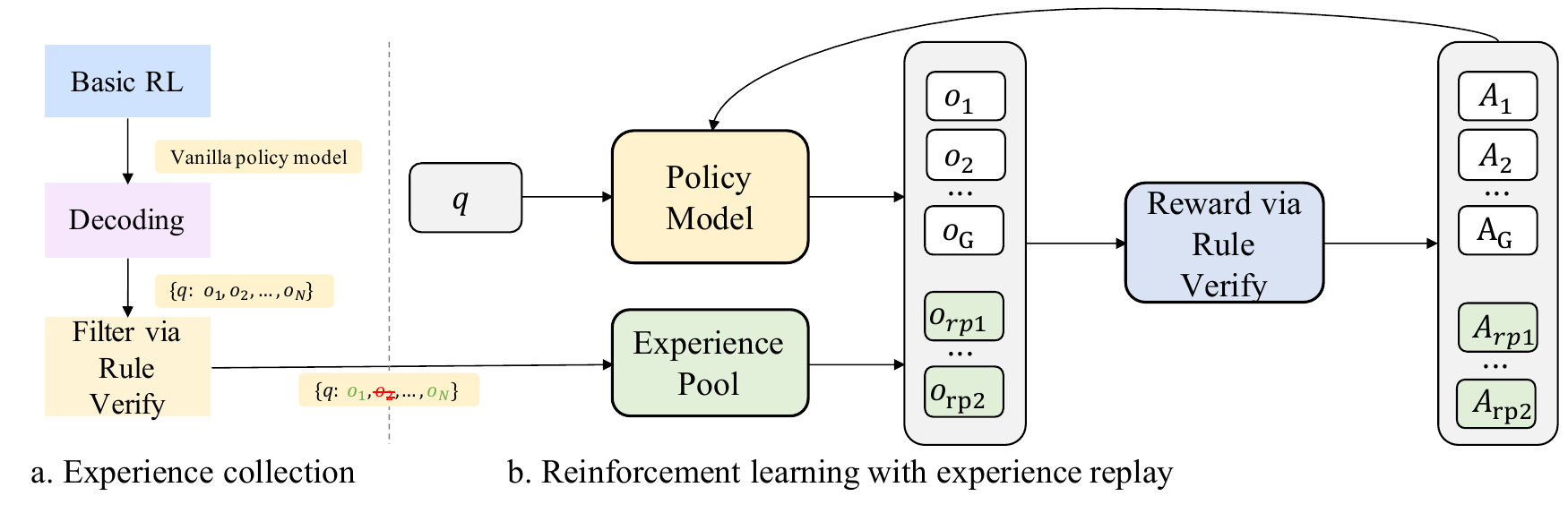}
    \centering
\caption{\textbf{RLEP} training pipeline. 
\textbf{(a) Experience collection.} After a preliminary vanilla RL run, the seed policy decodes multiple trajectories for each problem; those that reach a verified correct answer are retained and stored in an experience pool. 
\textbf{(b) Replay training.} At every update the current policy rolls out \(G\) fresh trajectories (blue). We then sample \(M\) successful trajectories from the experience pool (green) and merge them, yielding an enlarged batch of \(G' = G + M\). Advantages are computed over this mixed set.}

    \label{fig:rlep_method}
\end{figure}
The complete RLEP workflow—comprising the two stages of \emph{experience collection} and \emph{replay training}—is illustrated in Figure~\ref{fig:rlep_method}.

\paragraph{Experience Collection}\label{sec:exp_pool}
As illustrated in Figure~\ref{fig:rlep_method}a, we begin with a conventional RL run to obtain a seed policy. This policy decodes multiple candidate trajectories for every question, and any trajectory that reaches a verified correct answer is retained and stored in an \emph{experience pool} for later replay. 
For every question we maintain an \emph{experience pool}—a set of
trajectories that reach the correct answer, collected from previous RL‑trained checkpoints.

\paragraph{RLEP Training.}

Figure~\ref{fig:rlep_method}.b gives an overview of the RLEP training workflow. During each training update, we mix fresh roll‑outs with replayed successes as follows.

\begin{enumerate}[label=(\roman*)]
  \item \textbf{Rollout.}  
        The current policy $\pi_{\theta}$ generates a group of
        $G$ candidate trajectories, as in standard GRPO.

  \item \textbf{Experience replay.}  
        We randomly sample $M$ successful trajectories
        from the experience pool and append them to the freshly
        generated rollouts, expanding the group size from
        $G$ to $G' = G + M$.

  \item \textbf{Policy update.}  
  
Given the enlarged set of trajectories
$\{o_i\}_{i=1}^{G'}$ ($G' = G+M$), we apply the same
token‑level, asymmetrically clipped GRPO rule, now computed over the
\emph{mixed} group of fresh roll‑outs and replayed successes:

\begin{equation}
\begin{aligned}
\mathcal{J}_{\text{RLEP}}(\theta)=
\mathbb{E}_{\substack{q \sim P(Q),\\ \{o_i\}_{i=1}^{\textcolor{red}{G'}}\sim\pi_{\theta_{\text{old}}}(O\,|\,q)}}
\left[
\frac{1}{\sum_{i=1}^{\textcolor{red}{G'}} |o_i|}
\sum_{i=1}^{\textcolor{red}{G'}}
\sum_{t=1}^{|o_i|}
\min\left(
r_{i,t}(\theta)\,{A}_{i,t},\;
\operatorname{clip}\left(
r_{i,t}(\theta),\;1{-}\varepsilon_{\text{low}},\;1{+}\varepsilon_{\text{high}}
\right){A}_{i,t}
\right)
\right]
\end{aligned}
\label{eq:rlep_objective}
\end{equation}

\noindent
where $r_{i,t}(\theta)$ is the token‑wise importance ratio
$\pi_{\theta}(o_{i,t}\,|\,q)\big/\pi_{\theta_{\text{old}}}(o_{i,t}\,|\,q)$.
The advantage term $A_{i,t}$ is standardised over \emph{all}
$G'$ trajectories so that replayed successes and new roll‑outs
share a common baseline:

\begin{equation}
A_{i,t}
=\frac{r_{i,t}-\operatorname{mean}\!\bigl(\{r_{1,t},\dots,r_{G',t}\}\bigr)}
       {\operatorname{std}\!\bigl(\{r_{1,t},\dots,r_{G',t}\}\bigr)}.
\label{eq:rlep_advantage}
\end{equation}

Equations~\eqref{eq:rlep_objective} and~\eqref{eq:rlep_advantage}
thus generalise the GRPO update to the RLEP setting, ensuring that
every update step jointly leverages fresh roll‑outs
and  (replayed high‑quality trajectories.
\end{enumerate}

Replaying curated successful trajectories shields the policy from unproductive exploration, concentrates learning on fruitful lines of reasoning, and ultimately yields both quicker convergence and the potential to higher final accuracy.

%% file: sections/4-experiments.tex
\section{Experiment}
\label{sec:experiments}

\subsection{The Optimized Baseline}
\paragraph{Training strategy}
Starting from the hyper‑parameter recommendations in DAPO, we make several light adjustments and obtain consistent gains on AIME‑2024, AIME‑2025 and other datasets. Specifically, we adopt the \emph{token‑mean},   \emph{clip‑higher}  and  \emph{overlong-reward-shaping} strategies, while keeping most of the default settings in {Verl} \citep{sheng2024hybridflow}. 
Considering rollouts dominate the wall‑clock cost, we intentionally omit the \emph{dynamic‑sample} acceleration scheme. Instead, we fine‑tune the remaining hyper‑parameters to build a stronger baseline.

With 512 samples per rollout, the original configuration performs 16 actor updates using 32‑sample mini‑batches. 
Although this setting converges quickly, we observe a late‑stage decline in both BoN and Maj@N accuracy. Increasing the mini‑batch size to 64—i.e., eight updates per rollout—substantially improves training stability.

\begin{table}[h]
\small                
\centering
\begin{tabular}{l c c c}
\toprule
\textbf{Method} & \emph{token‑mean}, \emph{clip‑higher}, \emph{overlong‑shaping} & \emph{dynamic-sampling} & ppo\_mini\_batch\_size  \\
\midrule
DAPO             & \cmark & \cmark & 32 \\
DAPO‑nodyn‑bs32   & \cmark & \xmark     & 32 \\
DAPO‑nodyn‑bs64    & \cmark & \xmark     & 64 \\
\bottomrule
\end{tabular}
\caption{The tuned baseline configurations based on DAPO.}
\end{table}

\paragraph{Implement details.}  
We utilize Qwen2.5-Math-7B~\citep{yang2024qwen25mathtechnicalreportmathematical} as the base model, the input context length is set to 1024 wile the maximum response length is set to 3072. Each rollout stage takes 512 prompts. The temperature and top-$p$  value are set to 1.0 during training and validation. 

\begin{figure}[H]
    \centering
    \includegraphics[width=1\linewidth]{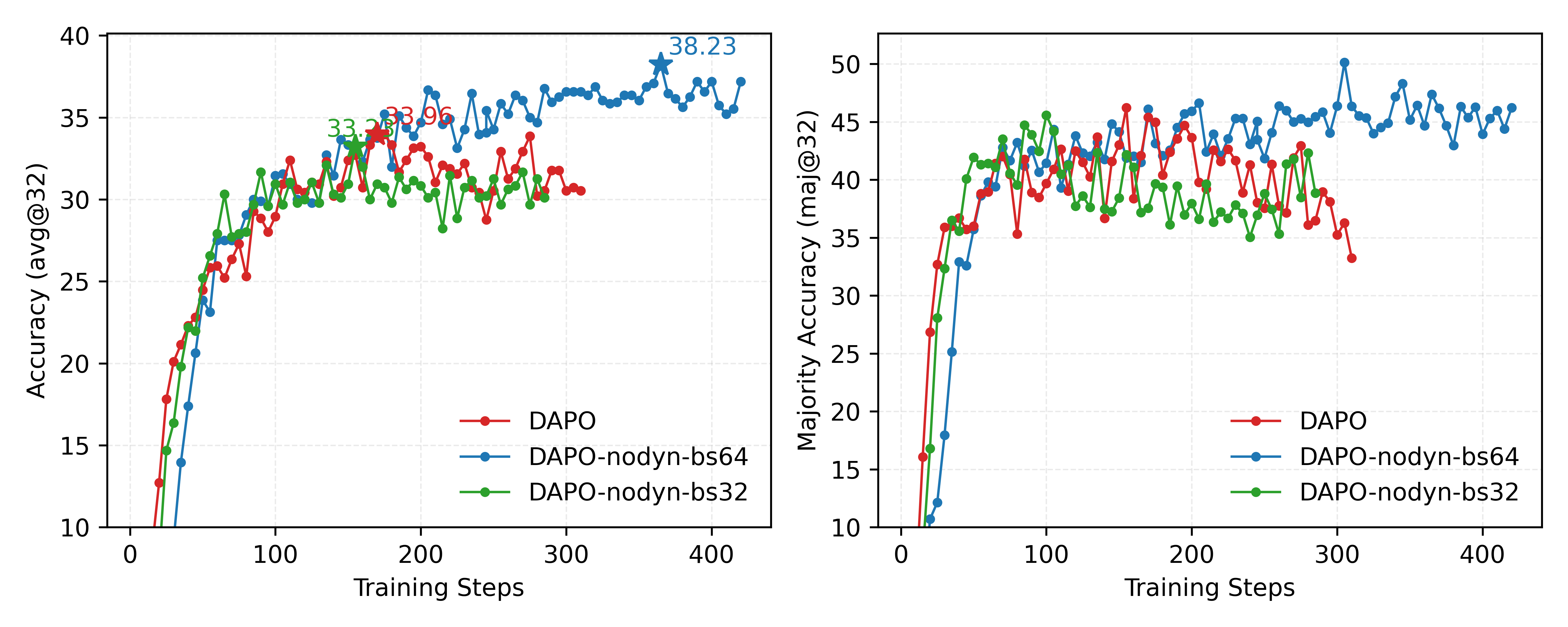}
    \caption{Performance of the optimized baseline method.}
    \label{fig:baseline_exp}
\end{figure}

Figure~\ref{fig:baseline_exp} compares several variants:
\begin{itemize}
    \item DAPO vs. DAPO-nodyn-bs32. 
    We find that DAPO with dynmic sampling achieves higher accuracy, show that the positive impact of dynamic sampling. 
    \item Comparing DAPO-nodyn with different PPO training mini-batchsize, the 32‑sample mini‑batch learns faster at the beginning, but the 64‑sample mini‑batch ultimately converges to higher accuracy and a smoother Maj@32 curve. The DAPO-nodyn-bs64 even slightly surpasses DAPO in overall accuracy, eliminating the impact of removing of dynamic sample.
    In practice, each DAPO update takes roughly 220 s before step 230, whereas DAPO‑nodyn‑bs64 needs only about 160 s. After step 230, DAPO’s per‑step time climbs to around 360 s because additional rollouts are required to fill the batch. Balancing speed and accuracy, we therefore use the configure DAPO‑nodyn‑bs64 for subsequent RLEP experiments.
    \item Model accuracy climbs quickly at the start of training, but as instability accumulates and the policy drifts from its initial weights, overall accuracy eventually plateaus—and can even decline—after a certain number of steps. This phenomenon shows that RL training is an energy‑intensive journey.

\end{itemize}

Typically, batch size has only a modest effect in standard supervised fine‑tuning (SFT). In reinforcement learning, each rollout is followed by several policy‑update steps, and batch size directly influences the proportion of samples affected by the advantage clipping operation. This coupling appears to be the reason batch size matters much more in RL.

\subsection{RLEP Experimental Results}

We start from the DAPO‑nodyn‑bs64 baseline, which trains for 400 PPO steps with a mini‑batch size of 64 to build the experience pool. For every question, the policy samples 64 candidate answers (temperature 0.7, top‑p 0.95); only answers verified as correct by the reward model are kept, and we require at least two such valid reasoning paths per question.
During the RLEP phase, each question receives 16 fresh on‑policy rollouts plus 2 replayed answers, while all other hyper‑parameters remain identical to the baseline.
Per‑step runtime increases by under 5 s relative to the DAPO‑nodyn‑bs64 baseline, leaving overall training time essentially unchanged.

\begin{figure}[!h]
    \centering
    \includegraphics[width=1\linewidth]{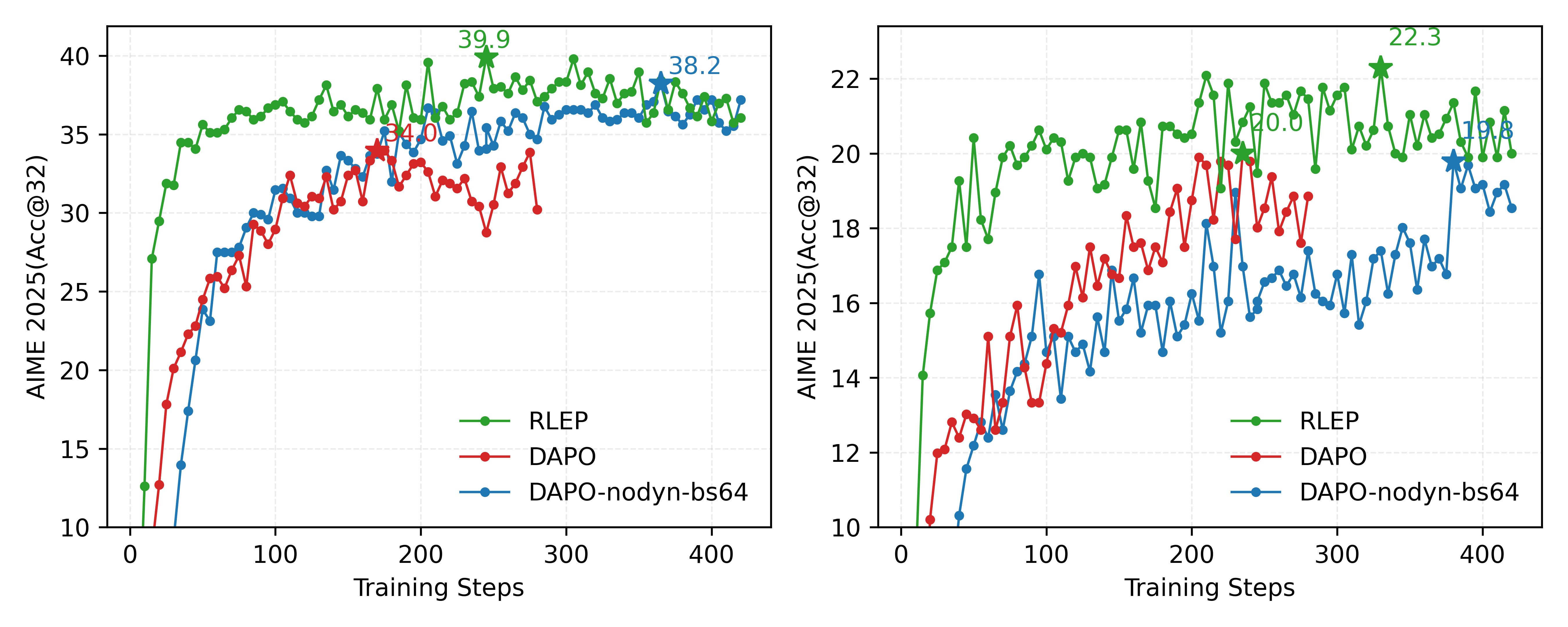}
    \caption{Main experimental results of RL wit Experience Replay.  }
    \label{fig:rlep_exp}
\end{figure}
The main results are shown in Figure~\ref{fig:rlep_exp} and can be summarised as follows:
\begin{itemize}
    \item \textbf{Rapid early gains.} With replayed experience, accuracy rises sharply at the start of training.  
    On the AIME‑2024 dataset, RLEP matches the baseline’s peak performance by step 135 (the baseline needs 380 steps).  
    On AIME‑2025, it surpasses the baseline’s best score after only 50 steps.  
    The replayed trajectories steer the model away from unproductive early exploration and difficult reasoning paths.
    \item \textbf{Higher final performance.} RLEP does more than accelerate convergence—it finishes higher.  
    The best accuracy on AIME‑2024 improves from 38.2\% to 39.9\%, and on AIME‑2025 from 19.8\% to 22.3\%. 
    (These points are peak value of the line, but similar conclusion could be drawn from the whole training procedure. )
    Evaluated offline on the unseen AMC‑2023 dataset, accuracy rises from 77.0\% to 82.2\%.  
    These results show that leveraging prior experience enables RLEP to converge to superior solutions.
    
\end{itemize}

Additionally, we examined whether supplementing the replay buffer with failed answers could help the policy avoid poor solutions. Replaying both successful and unsuccessful trajectories, however, produced no measurable improvement over positive replay alone. Error patterns vary widely across models and training stages—the mistake space is simply too broad—so unlikelihood updates on these heterogeneous errors provide little benefit to the current policy.

%% file: sections/5-conclusion.tex
\section{Conclusion and Future Work}
\label{sec:conclusion}

In this technical report we introduce Reinforcement Learning with Experience rePlay (RLEP).
During the experience‑collection phase, earlier RL runs act as pathfinders, tracing high‑reward trajectories. In the subsequent replay stage, the model rapidly converges to these previously discovered optima and then pushes beyond them to reach even stronger performance. Experimental results on AIME 2024, AIME 2025 and AMC 2023 demonstrates the effectiveness of RLEP.

In future work, we will further explore: (1) devising smarter experience‑selection schemes that leverage offline heuristics and model‑based rewards to identify the most informative reasoning paths for replay, and (2) extending RLEP beyond the single‐dataset setting by training on much larger corpora and evaluating its effectiveness across different domains.

%% file: sections/appendix.tex
\appendix